\documentclass{tlp}
\usepackage{mathptmx}

\usepackage{enumitem}
\usepackage{times}
\usepackage{helvet}
\usepackage{courier}
\usepackage{mathrsfs}
\usepackage[mathscr]{euscript}
\usepackage[all]{xy}
\usepackage{xparse}
\usepackage{verbatim}

\usepackage{csquotes}
\usepackage{lipsum}

\usepackage{color}
\usepackage{amsmath,amssymb}
\usepackage{latexsym}

\usepackage{xspace}
\usepackage{graphicx}
\usepackage{url}
\newcommand{\ignore}[1]{}

\usepackage{natbib}

\newtheorem{example}{Example}

\newcommand{\gcomp}{\mathit{Gcomp}}

\newcommand{\rul}{\leftarrow}
\renewcommand{\ar}{\leftarrow}
\newcommand{\naf}{\ensuremath{\mathtt{not}\;}}

\newcommand{\x}{\bar{x}}

\ignore{

es associated to an ENF sentence

}

\newcommand{\cred}[1]{{\color{black}#1}}

\newcommand{\gen}{{\em generate}\xspace}
\newcommand{\define}{{\em define}\xspace}
\newcommand{\test}{{\em test}\xspace}
\newcommand{\gdt}{\gen-\define-\test}

\def\beq{\begin{equation}}
\def\eeq#1{\label{#1}\end{equation}}

\def\citeay#1{\citeauthor{#1}~\citeyearpar{#1}}

\newcommand{\II}{{\mathbb I }} 

\NewDocumentCommand{\formulas}{g}{\IfValueTF{#1}{{\mathbb{L}_{#1}}}{{\mathbb{L}}}}
\NewDocumentCommand{\structures}{g}{\IfValueTF{#1}{{\mathbb{S}_{#1}}}{{\mathbb{S}}}}
\NewDocumentCommand{\infsemFO}{mgg}{
\IfValueTF{#2}{\mathscr{FO}_{#1}^{#2}}{\mathscr{FO}_{#1}}
\IfValueTF{#3}{(#3)}{}
}
\NewDocumentCommand{\infsemFOA}{mgg}{
\IfValueTF{#2}{\hat{\mathscr{FO}}_{#1}^{#2}}{\hat{\mathscr{FO}}_{#1}}
\IfValueTF{#3}{(#3)}{}
}
\NewDocumentCommand{\infsem}{mgg}{
\IfValueTF{#2}{\mathscr{I}_{#1}^{#2}}{\mathscr{I}_{#1}}
\IfValueTF{#3}{(#3)}{}
}
\NewDocumentCommand{\infsemGL}{mgg}{
\IfValueTF{#2}{\mathscr{GL}_{#1}^{#2}}{\mathscr{GL}_{#1}}
\IfValueTF{#3}{(#3)}{}
}
\NewDocumentCommand{\infsemGLorig}{mgg}{
\IfValueTF{#2}{\mathscr{G}_{#1}^{#2}}{\mathscr{G}_{#1}}
\IfValueTF{#3}{(#3)}{}
}

\NewDocumentCommand{\infsemDV}{mgg}{
\IfValueTF{#2}{\mathscr{DV}_{#1}^{#2}}{\mathscr{DV}_{#1}}
\IfValueTF{#3}{(#3)}{}
}

\NewDocumentCommand{\infsemGK}{mgg}{
\IfValueTF{#2}{\mathscr{GK}_{#1}^{#2}}{\mathscr{GK}_{#1}}
\IfValueTF{#3}{(#3)}{}
}

\NewDocumentCommand{\infsemT}{mgg}{
\IfValueTF{#2}{\mathscr{OB}_{#1}^{#2}}{\mathscr{OB}_{#1}}
\IfValueTF{#3}{(#3)}{}
}

\def\aspp{ASP-Prolog\xspace}
\def\citeb#1{(\citeauthor{#1}, \citeyear{#1})}

\def\citebbb#1#2#3{(\citeauthor{#1}, \citeyear{#1}; \citeauthor{#2}, \citeyear{#2};
\citeauthor{#3}, \citeyear{#3})}

\def\shortcite#1{(\citeyear{#1})}
\newcounter{numquote}
\newenvironment{lquote}[1]{%

  \stepcounter{numquote}
  \expandafter\xdef\csname#1\endcsname{\thenumquote}%
  \quote 
  \addtolength{\leftskip}{5mm}{
  {\bf Quote} \thenumquote: \ignorespaces}}{\unskip\endquote }
  
\newcommand\quoteref[1]{\csname#1\endcsname}

\begin{document}
\lefttitle{Yuliya Lierler}

\title[Historical Review of Variants of Informal Semantics for LP: GL'88, GL'91 and GK'14]{
Historical Review of Variants of Informal Semantics for Logic  Programs under Answer Set Semantics:\\ GL'88, GL'91, GK'14, D-V'12}

  \begin{authgrp}
\author{\gn{Yuliya Lierler}}
\affiliation{University of Nebraska Omaha}
\end{authgrp}
\jnlPage{\pageref{firstpage}}{\pageref{lastpage}}
\jnlDoiYr{2021}
\doival{10.1017/xxxxx}

\maketitle

\label{firstpage}
\begin{abstract}
This  note presents a historical survey of informal semantics that are associated with logic programming under answer set semantics. We review these  in uniform terms and align them with two paradigms: Answer Set Programming and \aspp --- two prominent Knowledge Representation and Reasoning Paradigms in Artificial Intelligence. Under consideration in Theory and Practice of Logic Programming (TPLP).
\end{abstract}

\maketitle

\section{Introduction}
The transcript of the talk by Donald E. Knuth titled {\em Let's Not Dumb Down the History of Computer Science} published by ACM~\shortcite{knu21} includes the statement:
\begin{quote}
\addtolength{\leftskip}{5mm}{
... it would really be desirable if there were hundreds of papers on history written by computer scientists about computer science.}
\end{quote}
This quote was inspirational  for this technical note devoted to a historical survey of informal semantics that are associated with logic programming under answer set semantics (in the sequel we mostly drop {\em under answer set semantics} when referring to logic programming and logic programs). 

We focus on four seminal publications  and  align informal semantics discussed there using the same style of presentation and propositional programs. We trust that within such  settings key ideas and tangible differences between the distinct views come to the surface best. The earliest publication of the four dates back to 1988, and the latest dates back to 2014. It would seem that the subject of informal semantics is only peripheral scoring at such a low count of major references. Rather, the word {\em informal}  makes this subject rare in the discussions of logic programming. Nevertheless, the 2014 reference is an \underline{introductory} chapter titled {\em Informal Semantics}  of the textbook on {\em Knowledge Representation, Reasoning, and Design of Intelligent Agents} by Gelfond and Kahl. The prominent position of this chapter  points at the importance of the subject, {\em especially when we consider passing on the knowledge and practice of logic programming to a broad audience}.

As the presentation unfolds, a story of two views on logic programs will emerge. One view is via the prism of answer set programming (ASP) and another view is via the prism of ASP-Prolog. We reserve the  term --- ASP  --- to a constraint programming paradigm, where an ASP practitioner
while coding specifications of a considered problem  ensures that the solutions to this problem correspond to answer sets of the coded program~\citeb{BrewkaET11}. 
ASP is frequently associated with solving difficult combinatorial search problems via a programming methodology of \gdt and underlying grounding and solving technology.
The term --- ASP-Prolog --- is used to denote a knowledge representation language geared to model and capture domain knowledge with the underlying intelligent/rational agent in mind~\cite[Section~2]{gelkahl14}.  One may utilize ASP-Prolog  as a programming language, but  may also simply use it for  describing specifications without thinking about a computational task or solving this task. 

This presentation of the four surveyed publications almost follows their timeline starting with the earliest work.  In many places, we present the original quotes from the discussed sources  to avoid misrepresentation of the originals.

				
\section{
				 Formal and informal semantics of basic  programs by GL'88}
We start by recalling the formal and informal semantics of basic logic programs as they were introduced by \citeay{ge1}.

A {\em basic rule}  is an expression of the form
				\beq A \rul B_1, \dots, B_n, \naf C_1, \dots,
				\naf C_m,
				\eeq{eq:rule}
				where $A$, $B_i$, \cred{and} $C_j$ are propositional atoms.
						The {atom} $A$ is the \emph{head} of the rule and the expression $B_1,\ldots,B_n, \naf C_1, \dots, \naf C_m$ is its \emph{body}. 
A \emph{basic (logic) program} is 
				a finite set of such rules. In the sequel, we introduce rules of somewhat different syntactic structure, yet we agree to call 
								the left-hand side of the rule operator/connective, denoted by $\rul$,  {\em head} and the right-hand side  {\em body}. A rule whose body is empty $(n=m=0)$ is called a {\em fact}; in such rules connective $\rul$ is often dropped.

For a rule
				$r$ of the form~\eqref{eq:rule} and a  set $X$ of atoms,  
				the reduct $r^X$  is defined whenever there is
				no atom $C_j$ for $j \in
				\{1,\ldots,m\}$ such that $C_j \in X$. If the reduct $r^X$ is defined, then it
				is the rule 
				\beq
				A \rul B_1,\ldots,B_n.
				\eeq{eq:rpos}
				The reduct $\Pi^X$ of the program $\Pi$ consists of the rules $r^X$ for all 
				$r\in \Pi$, for which the reduct is defined.
				A set $X$ of atoms {\em satisfies} rule~\eqref{eq:rpos} if $A$ belongs to $X$ or there exists $i\in\{1,\dots,n\}$ such that $B_i\not \in X$. We say that a set $X$ of atoms is a {\em model} of a  program consisting of rules of the form~\eqref{eq:rpos} when $X$ satisfies all  rules of this program.
				A set $X$ is a \emph{stable model}/\emph{answer set} of $\Pi$,
				denoted $X \models_{st} \Pi$, 
				if it is a subset minimal model of~$\Pi^X$. 

\paragraph{Quotes by \citeauthor{ge1}~\citeyearpar{ge1} on Intuitive Meaning of Basic Programs}
 (verbatim modulo names for programs and sets of atoms):
\begin{lquote}{q:1}
    The intuitive meaning of stable sets\footnote{\citeauthor{ge1}~\citeyearpar{ge1} use terms stable set and stable model interchangeably.} can be described in the same way as the intuition behind ``stable expansions'' in autoepistemic logic: they are ``possible sets of beliefs that a rational agent might hold''~\citep{Moore} given~$\Pi$ as his premises.  If $X$ is the set of (ground) atoms that I consider true, then any rule that has a subgoal $not\ C$ with~$C\in X$ is, from my point of view, useless; furthermore, any subgoal $not\ C$ with $C\not\in X$ is, from my point of view, trivial. Then I can simplify the premises $\Pi$ and replace them by $\Pi^X$. If $X$ happens to be precisely the set of atoms that logically follow from the simplified set of premises $\Pi^X$, then I am ``rational''.
\end{lquote}
Later,
\citeauthor{ge2}~\citeyearpar{ge2} say about a basic program the following: 
\begin{lquote}{q:2}
    A ``well-behaved'' program has exactly one stable model, and the answer that such a program is supposed to return for a ground query $A$ is {\em yes} or {\em no}, depending on whether $A$ belongs to the stable model or not. (The existence of several stable models indicates that the program has several possible interpretations). 
    \end{lquote}

The historical roots of stable model semantics for logic programs as a formal tool for model-theoretic declarative semantics of Prolog~\citeb{kow88a} 
are apparent in these quotes. The expectation is to consider a well-behaved program with a single stable model. Yet, the authors acknowledge the possibility of programs with several stable models that
{\em indicates that the program has several possible interpretations} or induces  several {\em possible sets of beliefs}. 
In this paper, it will prove of value to distinguish between the concepts  {\em possible interpretations}
and {\em possible sets of beliefs}. 
In 1988 these terms were used as synonyms. 
 It is convenient to imagine that the concept of possible interpretation stands behind what we characterize here as ASP, whereas the concept of possible sets of beliefs stands behind ASP-Prolog. 
We now review these frameworks prior to an attempt to formalize the presented quotes that will result in what we denote as an original informal semantics of basic programs.

\subsection{ASP and \aspp}
\paragraph{Answer Set Programming}
\citeauthor{mar99}~\citeyearpar{mar99} and \citeay{nie99} open a new era for stable model semantics by proposing the use of logic programs under this semantics  as constraint programming paradigm for modeling combinatorial search problems. This marks the birth of  {\em ASP}. Here is what the abstract of the paper by \citeauthor{mar99} says:

\begin{quote}
\addtolength{\leftskip}{5mm}{
    We demonstrate that inherent features of stable model semantics naturally lead to a logic programming system that offers an interesting alternative to more traditional logic programming\footnote{``more traditional logic programming'' refers to Prolog~\citeb{kow88a}}\dots 
    The proposed approach is based on the {\em interpretation of program clauses as constraints}. In this setting programs do not describe a single intended model, but a family of stable models.  These stable models encode solutions to the constraint satisfaction problem described by the program. 
    \dots We argue that the resulting logic programming system is well-attuned to problems in the class NP, has a well-defined domain of applications, and an emerging methodology of programming.
 }   
\end{quote}
In other words, \citeauthor{mar99} and \citeauthor{nie99} propose to see logic rules of the program as specifications of the constraints of a problem at hand. Logic programming is seen as  a provider of a general-purpose modeling language that
supports  solutions for search problems. Let us make these claims precise by
considering the notion of a search problem following the lines by Brewka et al.~(\citeyear{BrewkaET11}). A {\em search problem}~$P$ consists of a set of instances with each
{\em instance}~$I$ assigned a finite set~$S_P(I)$ of solutions. 
In the proposal by \citeauthor{mar99} and \citeauthor{nie99}, 
 to solve a
search problem~$P$, one constructs a logic program~$\Pi_P$ that captures
problem specifications so that when extended with facts~$F_I$
representing an instance~$I$ of the problem, the answer sets of
$\Pi_P\cup F_I$ are in one to one correspondence with members in~$S_P(I)$. In other words, answer sets of $\Pi_P\cup F_I$
describe all solutions of problem~$P$ for the instance
$I$. Thus, solving  a search problem $P$ is reduced to finding a uniform 
logic program -- that we denoted as~$\Pi_P$ -- which  encodes problem's  specifications/constraints. 

The logic rules of the program -- the key syntactic building blocks of logic programming -- become the vehicles for stating constraints/specifications of a problem under consideration.
A program is typically evaluated by means of a grounder-solver pair. A grounder~\citebbb{syr01}{geb07b}{cal08} is responsible for eliminating first order variables occurring in a logic program in favor of suitable object constants resulting in a propositional program -- atoms of such programs are called {\em ground}. An answer set solver -- a system in the spirit of SAT solvers (see, for example,~\citeb{lier16a}) -- is responsible for computing  answer sets (solutions) of a program. Let us draw a parallel with Prolog.
In Prolog~\citeb{kow88a} a {\em single} intended model is assigned to a logic program. The SLD-resolution~\citeb{kow74} is at the heart of the control mechanism behind Prolog implementations. Together with a logic program, a Prolog system expects a query (possibly multiple queries). This query is then evaluated by means of SLD-resolution and a given program against an intended model. 
Thus, even though Prolog and ASP share the basic language of logic programs, their  programming methodologies and underlying solving/control technologies are different.\footnote{
A term ``constraint logic programming''~\citeb{jaf87}
may come to reader's mind. This concept is vaguely related to the discussion here. It comes from an era predating logic programs under answer set semantics. Yet, it is also a form of
 constraint programming, in which logic programming -- understood as Prolog -- is extended to include concepts from constraint satisfaction~\citeb{dech03}. For instance,  a program may contain numeric constraints in the bodies of its rules. In some way, constraint logic programming is closer in spirit to what is called constraint answer set programming (see, for example,~\citeb{LIERLER20141}), where a logic program under answer set semantics may, for instance, contain numeric constraints in the bodies of its rules.}

\paragraph{ASP Programming Methodology}
\citeauthor{eit00}~\citeyearpar{eit00} illustrate how logic programs under answer set semantics
can be used to encode problems in a highly declarative fashion, following a ``{\em Guess\&Check}'' paradigm. 
We restate this paradigm verbatim utilizing the terminology of search problem and its instance introduced before.
\begin{quote}
\addtolength{\leftskip}{5mm}{
 Given a set~$F_I$ of facts that specify an instance~$I$ of some problem $P$, a {\bf Guess\&Check} program $\Pi$ for~$P$ consists of the following two parts

 {\bf Guessing Part}: The guessing part $G\subseteq \Pi$ of the program defines the search space, in a way such that answer sets of $G\cup F_I$ represent ``solution candidates'' for $I$.

{\bf Checking Part}: The checking part $C\subseteq \Pi$ 
of the program tests whether a solution candidate is in fact a solution, such that the answer sets 
$G\cup C \cup F_I$ 
represent the solutions of the problem instance $I$.    }
\end{quote}
 \citeauthor{eit00} point at a close relation of the {\em Guess\&Check} approach with the generate-and-test paradigm in the AI community~\citeb{win92}.

\citeauthor{lif02}~\citeyearpar{lif02} refines the {\em Guess\&Check} approach and  coins a term {\gdt} for this emerging methodology in logic programming that splits program rules into three groups:
\begin{itemize}[topsep=0pt,itemsep=-1ex,partopsep=1ex,parsep=1ex]
\item the \gen group is responsible for defining a large class of ``potential solutions''; 
\item the \test group is responsible for stating conditions to weed out potential solutions that do not  satisfy the problem's specifications; and 
\item the \define group is responsible for defining concepts that are essential in stating the conditions of \gen and \test. 
\end{itemize}
In this work, ``the idea of ASP is to represent a given computational problem by a program whose answer sets correspond to solutions, and then use an answer set solver to find a solution''. The paper illustrates the use of ASP to solve a sample planning problem. Yet, there are no  references to how one would intuitively read, for example, an occurrence of atom $A$ or expression $not\ A$ in rules. This is also true of many other instances of papers describing various applications of ASP.

To the best of our knowledge~\citeauthor{den12}~\citeyearpar{den12} in addition to earlier reviewed quotes in this section are the  two major accounts for reconciling the use of ASP (not \aspp) in practice and intuitive readings of answer sets and rules of programs. The previous to the last section reviews an account by~\citeauthor{den12}, while the remainder of this section predominantly concerns making already reviewed quotes more precise.

\paragraph{\aspp}
We reviewed the concept of answer set programming as championed by~\cite{mar99,nie99}.
In this view, a search problem at hand is a center piece. An ability to model a considered search problem by means of a logic program  so that the answer sets of this program are in one-to-one correspondence to the problem's solutions constitutes this paradigm.

The textbook by \cite{gelkahl14} focuses on an alternative practice of logic programming under answer set semantics.
It champions a view  on answer sets as possible sets of beliefs. 
 Interpreting  answer sets as such implies the presence of an intelligent agent behind a program.
 The program itself is seen to represent a knowledge base of this agent.  
 This corresponds to the view of a logic program  as a knowledge representation and reasoning  formalism {\em for the design of intelligent agents}. This is largely a position advocated in the  \citeauthor{gelkahl14} textbook. 
 We use the term \aspp to denote such use of logic programs. It makes sense to reflect on the notion of an intelligent agent provided in Section 1.1 of the cited textbook:
\begin{quote}
\addtolength{\leftskip}{5mm}{
In this book when we talk about an {\bf agent}, we mean an entity that observes and acts on an environment and directs its activity toward achieving goals. Note that this definition allows us to view the simplest programs as agents. A program usually gets information from the outside world, performs an appointed reasoning task, and acts on the outside world, say by printing the output, making the next chess move, starting a car, or giving advice. If the reasoning tasks that an agent performs are complex and lead to nontrivial behavior, we call it intelligent.
}
\end{quote}

Brief discussions by~\cite{ge2} and  Section 2.2.1 of the mentioned textbook  are major two accounts that speak of 
 intuitive readings of answer sets and rules of programs
when  \aspp is used in practice. Sections~\ref{sec:infsemGL} and~\ref{sec:infsemGL2} of this note are devoted to these accounts.

\subsection{``Formalizing'' Quotes~\quoteref{q:1} and~\quoteref{q:2} or Informal Semantics of Basic  Programs by GL'88}\label{sec:inf1}

Before we attempt to make the claims of  Quote~\quoteref{q:1}  
by \citeauthor{ge1}~\citeyearpar{ge1} precise it is due to discuss three interrelated and yet different concepts and how we understand them within this note: 
\begin{itemize}
    \item a state of affairs,
    \item a belief state, and
    \item a set of beliefs/belief set.
\end{itemize}
The following example is our key vehicle in this discussion. 

\begin{example}\label{ex:0}
	Consider a toy world with four possible {\em states of affairs} that fully describe it:
 \smallskip
 
 \begin{tabular}{ll|ll}
 1& Mary is a student& 2& Mary is a student\\
 & John is a student& & John is not a student\\
\hline
 3& Mary is not a student&  4& Mary is not a student\\
 & John is a student&
 & John is not a student\\
\end{tabular}
\smallskip

A {\em belief state} is associated with/represented by a conglomeration of states of affairs. In other words, a belief state assumes that multiple states of affairs can be deemed as possible by an  agent. Thus, a belief state is often associated with an agent who has  partial knowledge of the world.

Returning to our toy world,
the powerset of the listed four states of affairs forms the set of belief states for an agent operating in this world. For example,
if an agent assumes a belief state consisting of states 1, 2, 3, 4 of affairs -- let us denote it as $bs_1$--, we may conclude that this agent deems everything possible (or knows nothing factual about the world).
If an agent assumes a belief state consisting of states 1 and 2 of affairs -- let us denote it as $bs_2$--, we may conclude that this agent is aware of the fact that {\em Mary is a student}, whereas {\em John may or may not be a student}. In turn, if an agent assumes a belief state consisting of a single state 1 -- let us denote it as $bs_3$--, then we may conclude that this agent is aware of the two facts: {\em Mary is a student} and {\em John is a student}.

We now connect the concepts of a belief set (or, a set of beliefs) and  a belief state. The former is an abstraction of the latter. In other words, we understand belief sets as entities that capture/encode belief states. This encoding may lose some information so that multiple belief states may be ``consistent'' with a single belief set. For instance, in  the context of our toy world, a belief set consisting of a single proposition {\em Mary is a student} is consistent with any belief state that contains either state 1 of affairs or state 2 of affairs. Note how this belief set cannot distinguish between these different belief states. Indeed, a belief set consisting of a single proposition {\em Mary is a student} cannot distinguish between belief states $bs_1$, $bs_2$, and $bs_3$.
\end{example}

We are now ready to return to the claims of  Quote~\quoteref{q:1}  
by \citeauthor{ge1}~\citeyearpar{ge1} and attempt to make these
precise {\em with the allowance} that programs with multiple answer sets that   correspond to possible sets of beliefs/possible interpretations
are as valid programs as so-called well-behaved programs. 
%
In other words, 
per Quote~\quoteref{q:1} each answer set represents a set of beliefs of a rational agent (or $I$); 
thus this agent may have multiple sets of beliefs.
In the sequel, we drop the reference to ``or I'' and use ``an agent'' in the discourse\footnote{In personal communication with Michael Gelfond on the 27th of April, 2023, he confirmed that ``I'' in Quote~\quoteref{q:1} was meant to refer to a rational agent invoked in the same quote.}. 
In the case of basic programs, we take 
an understanding that  
\beq
\hbox{{\em the absence of an atom $A$ in a stable model represents the fact that  $A$ is false}.} 
\eeq{eq:imp}
This understanding is consistent with the view by \citeauthor{ge1}, which is reiterated in Quote~4 
in Section~\ref{sec:infsemGL}. 

Claim~\eqref{eq:imp} on the interpretation of answer sets has profound ramifications. Namely, this claim makes the three concepts ---
 a state of affairs,
a belief state, and
 a set of beliefs/belief set
--- exemplified earlier by highlighting their differences collapse into a single entity.
Let us use an example to illustrate this point.
    \begin{example}\label{ex:00}
	Recall the toy world from Example~\ref{ex:0}. Let us take atoms
 $student(mary)$ and $student(john)$ to represent propositions {\em Mary is a student} and {\em John is a student}, respectively. If our signature of discourse is composed only of these two atoms, we can construct four distinct subsets of atoms within this signature, namely, 
\beq
 \{student(mary),~~student(john)\};~~
 \{student(mary)\};~~
 \{student(john)\};~~ \emptyset.
 \eeq{eq:samplesets}
 Each of these sets of atoms can be identified in a natural manner with one of the four states of affairs that capture our toy world from 
 Example~\ref{ex:0}\footnote{It is common in propositional logic to identify sets of atoms over particular signature with interpretations: namely, an atom that is part of the set is considered mapped to {\em true} in the respective interpretation, whereas an atom not in the considered set is mapped to {\em false}.}; below we rewrite the table presented in that example by substituting propositions depicting distinct states of affairs with the respective sets of atoms:

 \smallskip
 
 \begin{tabular}{ll|ll}
 1& $\{student(mary)$,& 2& $\{student(mary)\}$\\
 &  $student(john)\}$& & \\
\hline
 3& &  4& $\emptyset$\\
 & $\{student(john)\}$&
 & \\
\end{tabular}
\smallskip

 Alternatively, let us take sets listed in~\eqref{eq:samplesets} to represent distinct belief states under the assumption that any atom not listed explicitly in a set under consideration is considered to be {\em false}. 
 Thus, the belief state $\{student(mary),~~student(john)\}$ represents a belief state consisting of a single  state of affairs denoted by 1 and represented by the same set of atoms as illustrated above;
 belief state $\{student(mary)\}$ represents a belief state consisting of
  a single  state of affairs denoted by  2 and represented by the same set of atoms.
 The same observation holds for the remaining two belief states depicted by sets of atoms in~\eqref{eq:samplesets}.
Thus, given the considered settings we may identify belief states and states of affairs.
The same argument applies to the concept of a belief set.
\end{example}



We denote  the informal semantics 
for basic programs
as~$\infsemGLorig{\II}$,  where $\II$ stands for {\em intended interpretations} of the program's  propositional atoms. It is typical in the informal semantics for classical logic expressions that each atom $A$ has an intended interpretation, $\II(A)$, which is represented  linguistically as a noun phrase about the application domain.
The informal semantics~$\infsemGLorig{\II}$ consists of three components:
				\begin{itemize}[topsep=0pt,itemsep=-1ex,partopsep=1ex,parsep=1ex]
\item the interpretation of structures --- here, answer sets --- denoted by  $\infsemGLorig{\II}{\structures}$, 
\item 				the interpretation of syntactical expressions in a program, denoted
				by $\infsemGLorig{\II}{\formulas}$, and
\item the interpretation of the
semantic relation --- here,  satisfaction --- denoted by $\infsemGLorig{\II}{\models}$.
\end{itemize}
The first component determines
a function from an answer set/a set of beliefs encoded by a set $X$ of atoms to a belief state of some agent (or, a state of affairs). 
The second component determines the informal reading of syntactical expressions in a program.
The third component determines the informal reading of the satisfaction relation.

					In the view of informal semantics $\infsemGLorig{\II}$, an answer set encodes 
     {\em a} belief state of some agent (or, a state of affairs --- as we have seen earlier, these concepts are indistinguishable under claim~\eqref{eq:imp}). To reiterate,
 {\em 
					An agent in some belief state -- represented by a set $X$ of atoms -- considers the set of all atoms in~$X$ to be the case (believes in them), whereas any atom that does not belong to~$X$ is believed to be false by the agent, i.e., is not the case.}
					 Thus,  we may  explain the meaning of a program in terms of what atoms an agent with its knowledge  of the application domain encoded as the program  believes as true and what atoms an agent believes as false.
					 Generally, an agent in some belief state considers certain states of affairs as possible and others as impossible. 
					For basic programs, set~$X$ of atoms defines a {\em unique} state of affairs that the agent regards as possible 
					in a belief state that~$X$ represents.
					Thus, we may identify any belief state captured by~$X$
					with this state of affairs.
    We denote a state of affairs captured by a set $X$ of atoms  under an intended interpretation~$\II$ as $\infsemGLorig{\II}{\structures}{X}$. 					Table~\ref{fig:GLinf:struc1} summarizes 
    a role of an answer set as a state of affairs in the considered view.
			
				\begin{table}
					\caption{The Gelfond-Lifschitz  \citeyearpar{ge1}  informal semantics of answer sets -- sets of atoms.\label{fig:GLinf:struc1}}
					\begin{center}
						\begin{tabular}{ccl}
							\hline
							A set  $X$ of atoms& \phantom{aaaaa} &
				
							A state $\infsemGLorig{\II}{\structures}(X)$ of affairs \rule{0pt}{0.4cm}
							\\
							\hline\hline
							$A \in X$ for atom $A$ & & \begin{minipage}{6cm} $\II(A)$ is true  
							in state $\infsemGLorig{\II}{\structures}(X)$  of affairs  \end{minipage}\\
							\hline
							$A \not \in X$ for atom $A$ & & 							 \begin{minipage}{6cm}
								$\II(A)$ is false in state $\infsemGLorig{\II}{\structures}(X)$  of affairs \end{minipage}\\
							\hline\hline
						\end{tabular}
					\end{center}
				\end{table}

				\begin{example}\label{ex:1}
	Consider  a set of beliefs encoded as a set 
	\beq
	X=\{student(mary),~~male(john)\}
	\eeq{eq:ex1}
	of atoms 
	under the obvious intended interpretation $\II$ for the propositional atoms in $X$. This $X$ represents  a state of affairs in which the agent considers that both statements {\em Mary is a student} and  {\em John is a male} are true. At the same time, the agent considers any other statements, including {\em John is a student} and {\em Mary is a male}, false. The $\infsemGLorig{\II}{\structures}{X}$ component of informal semantics of basic programs provides us with this understanding of set $X$.
							\end{example}

				\begin{table}
					\caption{The Gelfond-Lifschitz \citeyearpar{ge1} informal semantics for basic logic programs.\label{fig:GLinf:form1}}
					\begin{center}
						\begin{tabular}{llcp{5.8cm}}
							\hline
							&\hspace{2cm}$\Phi$  & \phantom{} & \rule{0pt}{0.2cm}\hspace{2cm} $\infsemGLorig{\II}{\formulas}{\Phi}$\\
							\hline\hline
							1.~~& propositional atom $A$ & & { $\II(A)$ } \rule{0pt}{0.2cm}\\ 
							\hline			
							2.~~& expression of the form {\tt not~}$C$ & & it is not the case that $\II(C)$
							\rule{0pt}{0.2cm}\\ 
							\hline 
							3.~~& expression of the form $\Phi_1,\Phi_2$ & & $\infsemGLorig{\II}{\formulas}{\Phi_1}$ and $\infsemGLorig{\II}{\formulas}{\Phi_2}$ \rule{0pt}{0.2cm}\vspace{0.0cm}\\
							\hline
							4.~~& rule $Head\rul Body$ & & \begin{minipage}{5.8cm}
								if $\infsemGLorig{\II}{\formulas}{Body}$ then $\infsemGLorig{\II}{\formulas}{Head}$\\
								\emph{(in the sense of material implication)}
\end{minipage}\rule{0pt}{0.2cm}\vspace{0.0cm}\\
							\hline
							5.~~& program $P=\{r_1,\ldots,r_n\}$ & & \begin{minipage}{5cm}
								\rule{0pt}{0.2cm}  All the agent knows is:
								
								$\infsemGLorig{\II}{\formulas}{r_1}$ and $\infsemGLorig{\II}{\formulas}{r_2}$  and \dots
								$\infsemGLorig{\II}{\formulas}{r_n}$
							\end{minipage}\vspace{0.0cm}\\
							\hline\hline
						\end{tabular}
					\end{center}
				\end{table}

				Table \ref{fig:GLinf:form1} shows the Gelfond-Lifschitz  \citeyearpar{ge1}  informal semantics $\infsemGLorig{\II}{\formulas}$ of syntactic elements of programs.
        The term {\em material implication} used within the table 
        assumes the conformance to the usual truth table of implication and thus a conditional statement can be identified with  a disjunction in which the antecedent of the conditional statement is negated. 
								As it is clear from this table, under $\infsemGLorig{\II}{\formulas}$, extended logic programs have both classical and non-classical
				connectives\footnote{We understand connectives reminiscent of the ones appearing in classical propositional logic as classical.}. On the one hand, the comma connective (appearing in the body of rules) is classical conjunction and the rule
				connective $\rul$ is the classical implication. 
   Note that such reading of the comma connective and the rule connective allows us to identify the empty body of the rule with $\top$ --- a propositional constant whose value is always interpreted as {\em true} --- and  a rule with an empty body with a simple propositional statement that contains only head of this rule. Not surprisingly such rules are typically denoted as facts. 
    On the other hand, the implicit composition
				operator (constructing a program out of individual rules) is non-classical because it performs a closure operation (resulting in the implementation of closed world assumption ---  presumption that what is not currently known to be true is false):
				\emph{only} what is explicitly stated is known. 
    To summarize, Table \ref{fig:GLinf:form1}  is devoted to	the interpretation of syntactical expressions in a program allowing us to ``translate'' its syntactic elements  and the program itself into natural language expressions.
For instance, take $\II$ to be an identity function and consider a simple program\footnote{This  program stems from Example 2.2.1~\citeb{gelkahl14} and will reappear in Section~\ref{sec:infsemGL2}.	}
$$
	\begin{array}{ll}
	p(b)\ar q(a). ~~~~~~~~~~~~& \hbox{``If $q(a)$ then $p(b)$''}\\
	q(a). & \hbox{``$q(a)$''}\\
          &  \hbox{``The agents knows only the statements presented above''}\\
	\end{array}
	$$
The annotations to the right are warranted by  $\infsemGLorig{\II}{\formulas}$  presented in  Table \ref{fig:GLinf:form1}. 
				Table~\ref{fig:GLinf:models1} presents the final component $\infsemGLorig{\II}{\models_{st}}$ of the ~$\infsemGLorig{\II}$  informal semantics.
    In the context of the simple program used to illustrate  the findings presented in 
    Table \ref{fig:GLinf:form1}, the findings of Table~\ref{fig:GLinf:models1} suggest that the only answer set of this program $\{p(b),~~ q(a)\}$ is a state of affairs inferred from the knowledge encoded in this program  so that both $p(b)$ and $q(a)$ are the case and nothing else is the case.

				\begin{table}
					\caption{The Gelfond-Lifschitz \citeyearpar{ge1} informal semantics for the satisfaction relation.\label{fig:GLinf:models1}}
					\begin{tabular}{lcp{8.9cm}}
						\hline
						\hspace{0.5cm}$\models_{st}$ & \phantom{aaa}& \hspace{2cm}$\infsemGLorig{\II}{\models_{st}}$\\
						\hline\hline
						$X\models_{st} \Pi$ & & \rule{0pt}{0.4cm}Given  $\infsemGLorig{\II}{\formulas}{\Pi}$, $X$ could be a state of affairs 
						inferred from this knowledge
						so that 
						any proposition in $X$ is the case  whereas any proposition not in $X$ is not the case\\
						\hline\hline
					\end{tabular}
				\end{table}
				
\section{Formal and informal semantics of extended  programs by GL'91} \label{sec:infsemGL}
				Here, we recall the  formal and informal semantics of 
				extended logic programs by \citeay{ge2}. 
				An alternative view of the informal semantics for extended  logic programs is provided in~\cite[Section 2.2.1]{gelkahl14} reviewed next. 

				A \emph{literal} is either an atom $A$ or an expression $\neg A$, where $A$ is an atom. An {\em extended rule} is an expression of the form~\eqref{eq:rule},
				where $A$, $B_i$, \cred{and}~$C_j$ are propositional literals. 
				An \emph{extended program} is 
    a finite set of extended rules.  \citeay{ge2} also considered {\em disjunctive} rules   of the form
$D_1\ or\ \dots\ or\ D_l \rul B_1, \dots, B_n, \naf C_1, \dots,\naf C_m,$
where $D_k$, $B_i$, \cred{and} $C_j$ are propositional literals. Yet, the discussion of such rules is outside the scope of this note.

				A {\em consistent} set of propositional literals is  a set that does not contain
				both  $A$ and its {\em complement}~${\neg}A$ for any atom $A$.
				A \emph{believed literal set} $X$ is a consistent set of propositional literals. 
				A believed literal set $X$
				\emph{satisfies} an extended rule $r$ of the form~\eqref{eq:rule} if $A$ belongs to $X$ or
				there exists an $i \in \{1,\ldots,n\}$ such that $B_i \not\in X$ or a $j \in
				\{1,\ldots,m\}$ such that $C_j \in X$. A believed literal set is a {\em model} of a program $\Pi$ if
				it satisfies all rules in~$\Pi$.
				For a rule
				$r$ of the form~\eqref{eq:rule} and a believed literal set $X$,  
				the reduct $r^X$  is defined whenever there is
				no literal $C_j$ for $j \in
				\{1,\ldots,m\}$ such that $C_j \in X$. If the reduct $r^X$ is defined, then it
				is the extended rule of the form~\eqref{eq:rpos}.
								The reduct $\Pi^X$ of the program $\Pi$ consists of the rules $r^X$ for all 
				$r\in \Pi$, for which the reduct is defined.
				A believed literal set $X$ is an \emph{answer set} of $\Pi$,
				denoted $X \models_{st} \Pi$, 
				if it is a subset minimal model of $\Pi^X$. 
    (A subset minimal model is such that none of its strict subsets is also a model.)

	
	\paragraph{Quotes by \citeay{ge2} on Intuitive Meaning of Extended Programs}
\begin{lquote}{q3}
For an extended program, we will define when a set $X$ of ground {\em literals} qualifies as its {\em answer set}. \dots A ``well-behaved'' extended program has exactly one answer set, and this set is consistent. The answer that the program is supposed to return to a ground query $A$ is {\em yes}, {\em no}, or {\em unknown}, depending on whether the answer set contains $A$, $\neg A$, or neither. The answer {\em no} corresponds to the presence of explicit negative information in the program.
    
    Consider, for  instance, the extended program $\Pi_1$ consisting of just one rule:
$$\neg q\ar not\ p.$$

Intuitively, this rule means: ``$q$ is false, if there is no evidence that $p$ is true.'' We will see that the only answer set of this program is $\{\neg q\}$. The answers that the program should give to the queries $p$ and $q$ are, respectively {\em unknown} and {\em false}.

As another example, compare two programs that do not contain {\em not}:
$$
\neg p\ar, \ \ \ p\ar \neg q  \hbox{~~~~~~and~~~~~~}
\neg p\ar, \ \ \ q\ar \neg p
$$
... Thus our semantics is not ``contrapositive'' with respect to $\ar$ and $\neg$; it assigns different meanings to the rules $p\ar\neg q$ and $q\ar\neg p$. The reason is that it interprets expressions like these as {\em inference rules}, rather than conditionals.
\end{lquote}
This quote echos Quote~\quoteref{q:2} about basic programs: the notion of a well-behaved program resurfaces.
In comparison to basic programs, extended programs provide us with a new possibility to answer queries against a program --- namely,  {\em unknown}.
The following quote echos Quote~\quoteref{q:1} about basic programs:
\begin{quote}
\addtolength{\leftskip}{5mm}{
The answer sets of $\Pi$ are, intuitively, possible sets of beliefs that a rational agent may hold on the basis of information expressed by the rules of $\Pi$. 
If~$X$ is the set of (ground) literals that the agent believes to be true, then any rule that has a subgoal {\em not L} with $L\in X$ will be of no use to him, and he will view any subgoal {\em not L} with $L\not\in X$ as trivial. Thus he will be able to replace the set of rules $\Pi$ by the simplified set of rules~$\Pi^X$. If the answer set of $\Pi^X$ coincides with $X$, then the choice of $X$ as the set of beliefs is ``rational''.}
\end{quote}
The following quote states the precise relationship between basic and extended programs\footnote{In the original quote word basic was replaced by general, yet we use the terminology of this paper for clarity.}: 
\begin{lquote}{q:nk}
the semantics of extended programs applied to basic programs turns into the stable model semantics. But there is one essential difference: The absence of an atom $A$ in a stable model of a basic program represents the fact that $A$ is false; the absence of $A$ and $\neg A$ in an answer set of an extended program is taken to mean that nothing is known about $A$.
\end{lquote}

In the section on {\em Representing Knowledge Using Classical Negation}, \citeay{ge2} say 
\begin{quote}
\addtolength{\leftskip}{5mm}{
The difference between {\em not p} and $\neg p$ in a logic program is essential whenever we cannot assume that the available positive information about $p$ is complete, i.e., when the ``closed world assumption'' [Reiter, 1978] is not applicable to $p$. The closed world assumption for a predicate $p$ can be expressed in the language of extended programs by the rule 
$$
\neg p\ar \naf p 
$$
When this rule is included in the program, $not\ p$ and $\neg p$ can be used interchangeably in the bodies of other rules. Otherwise, we use {\em not p} to express that $p$ is not known to be true, and $\neg p$ to express that $p$ is false.}
\end{quote}
To summarize, \citeauthor{ge2} describe informal semantics for extended programs based on \emph{epistemic} notions of default and autoepistemic reasoning. We now present the informal semantics $\infsemGL{\II}$
for extended programs just as we presented $\infsemGLorig{\II}$ for basic programs. This presentation at times (in particular, Example~\ref{ex:2}) follows the lines by \citeay{den19}. 

\begin{table}
					\caption{The Gelfond-Lifschitz~\citeyearpar{ge2} informal semantics of answer sets -- sets of literals.\label{fig:GLinf:struc}}
					\begin{center}
						\begin{tabular}{ccl}
							\hline
							A believed literal set $X$& & A belief state $B \in\infsemGL{\II}{\structures}(X)$ that has abstraction $X$ \rule{0pt}{0.4cm}\\
							\hline\hline
							$A \in X$ for atom $A$ & & \begin{minipage}{7.4cm}$B$ has the belief that $\II(A)$ is true; i.e.,\\ $\II(A)$ is true in all states of affairs possible in $B$ \end{minipage}\\
							\hline
							$\lnot A \in X$ for atom $A$ & &  \begin{minipage}{7.4cm}
								$B$ has the belief that $\II(A)$ is false; i.e.,\\  $\II(A)$ is false in all states of affairs possible in $B$\end{minipage}\\
							\hline
							$A \not\in X$ for atom $A$ & &\begin{minipage}{7.4cm} $B$ does not have the belief that $\II(A)$ is 
								true; i.e.,\\  $\II(A)$ is false in some state of affairs possible in $B$\end{minipage}\\
							\hline
							$\lnot A \not\in X$ for atom $A$ & & \begin{minipage}{7.4cm}$B$ does not have the belief that $\II(A)$ is false; i.e.,\\ $\II(A)$ is true in some state of affairs possible in $B$\end{minipage}\\
							\hline\hline
						\end{tabular}
					\end{center}
				\end{table}

We begin by discussing  a crucial difference between $\infsemGLorig{\II}$ and $\infsemGL{\II}$.
					Informal semantics $\infsemGL{\II}$ 
					views a believed \underline{literal set} $X$ as an abstraction of a belief state (in fact, of a class of belief states that it cannot distinguish)  of some agent;  $\infsemGLorig{\II}$ views a  \underline{set $X$ of atoms} as  a state of affairs. The change from ``sets of atoms'' to ``sets of literals''
     and the elimination of  assumption~\eqref{eq:imp}
     are crucial. 
					Recall how an agent in some belief state considers certain states of affairs as possible and others as impossible. Within $\infsemGLorig{\II}$, set $X$ of atoms ends up representing a unique possible state of affairs
					associated with a belief state so that we may identify these two concepts. 
					Yet, believed literal set $X$ is the set of all literals $L$ that the agent believes in, that is, those that are true in all states of affairs that the agent regards as possible. Importantly, it is not the case that
						a literal $L$ that does not belong to $X$ is believed to be false by the agent.
					Rather, it is \emph{not believed} by the agent
					or as stated in Quote~\quoteref{q:nk} {\em nothing is known about $L$} to the agent. \citeay{den19} takes the following interpretation of a statement {\em literal $L$ is not believed by an agent/nothing is known about $L$}:
					  literal $L$ is false in some states of affairs the agent holds possible, and
					 $L$ must be true in at least one of the agent's possible states of affairs (unless the agent believes the complement of~$L$). 
					 This note adopts such an interpretation. 
					 We denote the class of informal belief states that are abstracted  to a given  formal believed literal set $X$ under an intended interpretation $\II$ as $\infsemGL{\II}{\structures}{X}$. Table \ref{fig:GLinf:struc} summarizes a view on a believed literal set as an abstraction of a belief state of some agent.


				\begin{example}\label{ex:2} We may view this example as a continuation of Example~\ref{ex:1}. Here we consider what would seem the same belief set but change the perspective on it from the point of view of informal semantics of basic programs to that of extended programs.
	Consider believed \underline{literal} set~\eqref{eq:ex1}
	under the obvious intended interpretation $\II$ for the elements in $X$. This $X$ is the abstraction of any belief state in which the agent believes that {\em Mary is a student} and  {\em John is a male}, and nothing is known about such statements as {\em John is a student} or  {\em Mary is a male}. One such belief state is the state $B_0$ in which the agent considers the following states of affairs as possible:
					\begin{enumerate}[topsep=0pt,itemsep=-1ex,partopsep=1ex,parsep=1ex]
						\item John is the only male in the domain of discourse; Mary is the only student.
						\item\label{it:2} John and Mary are both male students.
						\item\label{it:3} John and Mary are both male; Mary is the only student.
						\item\label{it:4} John is the only male; John and Mary are both students.
					\end{enumerate}
					Another belief state corresponding to $X$ is the state $B_1$ 
						in which the agent considers the states of affairs~\ref{it:2}-\ref{it:4} of $B_0$ as possible. 
						Indeed, for each of these belief states, it holds that Mary is a student and John is a male in all possible states of affairs of that belief state. Thus, each of the literals in $X$ is believed in each of the belief states $B_0$ and $B_1$. On the other hand, John is a student precisely in the state of affairs \ref{it:2} and \ref{it:4}; Mary is a male in the states of affairs \ref{it:2} and \ref{it:3}. Hence, literals $\neg student(john)$ and $\neg male(mary)$ are not believed in either of the two belief states $B_0$ and $B_1$.
							\end{example}


The component  $\infsemGL{\II}{\formulas}$ captures the informal readings of the connectives of the Gelfond-Lifschitz~\citeyearpar{ge2} informal semantics of extended programs. We summarize it by~(i)~the entries in rows 1, 3--5 of Table~\ref{fig:GLinf:form1}, where we replace $\infsemGLorig{\II}{\formulas}$ by $\infsemGL{\II}{\formulas}$, and~(ii)~the entries in
				Table \ref{fig:GLinf:form}.
				The definition of  $\infsemGL{\II}{\formulas}$ 
				suggests that of the two negation operators, symbol $\neg$ is classical negation, whereas {\tt not} is
				a non-classical negation. It is commonly called \emph{default negation}.
								The  component $\infsemGL{\II}{\models_{st}}$  explains what it means for a believed literal set $X$ 
				 to be an answer set/stable model of an extended program.  Table~\ref{fig:GLinf:models} presents its definition.

				\begin{table}
					\caption{The Gelfond-Lifschitz \citeyearpar{ge2} informal semantics for some expressions in extended programs.\label{fig:GLinf:form}}
					\begin{center}
						\begin{tabular}{lcp{5.8cm}}
							\hline
							\hspace{2cm}$\Phi$  & \phantom{aa} & \rule{0pt}{0.2cm}\hspace{2cm} $\infsemGL{\II}{\formulas}{\Phi}$\\
							\hline\hline
							propositional literal ${\neg}A$ & & it is not the case that $\II(A)$
							\rule{0pt}{0.2cm}\\ 
							\hline 
							expression of the form {\tt not~}$C$ & & the agent does not know that $\infsemGL{\II}{\formulas}{C}$\rule{0pt}{0.2cm}\vspace{0.0cm}\\ 
							\hline\hline
						\end{tabular}
					\end{center}
				\end{table}

				\begin{table}
					\caption{The Gelfond-Lifschitz \citeyearpar{ge2} informal semantics for the satisfaction relation.\label{fig:GLinf:models}}
					\begin{center}
					\begin{tabular}{lcp{9cm}}
						\hline
						\hspace{0.5cm}$\models_{st}$ & \phantom{a}& \hspace{2cm}$\infsemGL{\II}{\models_{st}}$\\
						\hline\hline
						$X\models_{st} \Pi$ & & \rule{0pt}{0.4cm}Given  $\infsemGL{\II}{\formulas}{\Pi}$,
					$X$ could be the 
						set of literals the agent believes\\
						\hline\hline
					\end{tabular}
					\end{center}
				\end{table}

We are now ready to comment on the meaning of querying a well-behaved extended program -- a program that has exactly one answer set.
Take $X$ to be the unique answer set/believed literal set of a well-behaved extended program.
In accordance with   Table~\ref{fig:GLinf:models}, $X$ is {\em the} set of literals the agent believes.
In accordance with  $\infsemGL{\II}$ summarized in Table \ref{fig:GLinf:struc}, set $X$ is an abstraction of a belief state $B \in\infsemGL{\II}{\structures}(X)$, where $B$ belongs to the unique class $C$ of belief states associated with~$X$. In turn, all members of $C$ are indistinguishable by their abstraction~$X$,  which characterizes some properties of possible states of affairs associated with all elements in $C$. Due to the uniqueness of the believed literal set, for the case of well-behaved extended programs, we may simplify the reading of the {\em unique} believed literal set $X$ as the abstraction of all possible states of affairs (from the perspective of a considered agent). In other words, what an agent believes coincides with factual information about the world. 
 Take an atom $A$ to be a query. The following table summarizes the interpretation of possible query responses.
 
				\begin{tabular}{ccl}
							\hline\hline       
       &query response&\\
       \hline\hline     
							$A \in X$& yes & \begin{minipage}{7.4cm}
       $\II(A)$ is true in all possible states of affairs  \end{minipage}\\
							\hline
       					$\lnot A \in X$  & no &   \begin{minipage}{7.4cm}
 $\II(A)$ is false in all possible states of affairs \end{minipage}\\
							\hline
    
							$A \not\in X$ and $\lnot A \not\in X$  & unknown &\begin{minipage}{7.4cm}   $\II(A)$ is false in some possible state of affairs  
        and
        $\II(A)$ is true in some other possible state of affairs\end{minipage}
                \\
							\hline\hline
						\end{tabular}

Provided account of informal semantics of extended logic programs
echos the interpretation of an answer set of an extended program as a possible ``set of beliefs'' and can be seen as informal semantics for the syntactic constructs that are fundamental in \aspp.

\section{Informal semantics of extended logic programs by GK'14} \label{sec:infsemGL2}	
\citeay{gelkahl14} consider a language of extended logic programs with the addition of
(i) disjunctive rules and (ii)
  rules called {\em constraints} that have the form
\beq  \rul B_1, \dots, B_n, \naf C_1, \dots,\naf C_m,
				\eeq{eq:con}
where $B_i$, \cred{and} $C_j$  are propositional  literals (empty head can be identified with $\bot$).
It is due to note that constraints have been in the prominent use of ASP/\aspp for some time. In particular, they are the kinds of rules that populate the \test group of \gdt programs mentioned earlier. We come back to this point in the next section.  
To generalize the concept of an answer set to extended programs with constraints it is sufficient to provide a definition of rule satisfaction when the head of the rule is empty: A believed literal set $X$ {\em satisfies} a constraint~\eqref{eq:con},  if there exists an $i \in \{1,\ldots,n\}$ such that $B_i \not\in X$ or a $j \in
				\{1,\ldots,m\}$ such that $C_j \in X$. As before, we do not  present definitions for programs with disjunctive rules.

\paragraph{Quote by \citeauthor{gelkahl14}~\citeyearpar{gelkahl14} on Intuitive Meaning of Extended Programs with Constraints}

	\begin{quote}
 \addtolength{\leftskip}{5mm}{
	    		Informally, program $\Pi$ can be viewed as a specification for answer sets –--
sets of beliefs that could be held by a rational reasoner associated with $\Pi$.
Answer sets are represented by collections of ground literals. In forming
such sets the reasoner must be guided by the following informal principles:
\begin{enumerate}[topsep=0pt,itemsep=-1ex,partopsep=1ex,parsep=1ex]
    \item  Satisfy the rules of $\Pi$. In other words, believe in the head of a rule if
you believe in its body.
\item  Do not believe in contradictions.
\item  Adhere to the “Rationality Principle” that says, “Believe nothing you
are not forced to believe.”
\end{enumerate}
Let’s look at some examples. \dots
	
	\smallskip
	\noindent
	{\bf Example 2.2.1.}
	$$
	\begin{array}{ll}
	p(b)\ar q(a). & \hbox{``Believe $p(b)$ if you believe $q(a)$''}\\
	q(a). & \hbox{``Believe $q(a)$''}\\
	\end{array}
	$$
	Note that the second rule is a fact. Its body is empty. Clearly, any set of literals
satisfies an empty collection, and hence, according to our first principle,
we must believe $q(a)$. The same principle applied to the first rule forces
us to believe $p(b)$. The resulting set $S1 = \{q(a), p(b)\}$ is consistent and
satisfies the rules of the program. Moreover, we had to believe in each of its
elements. Therefore, it is an answer set of our program. Now consider set
$S2 = \{q(a), p(b), q(b)\}$. It is consistent, and satisfies the rules of the program,
but contains the literal $q(b)$, which we were not forced to believe in by our
rules. Therefore, $S2$ is not an answer set of the program. \dots

\smallskip
	\noindent
	{\bf Example 2.2.2.} {\em (Classical Negation)}
	$$
		\begin{array}{ll}
	\neg p(b)\ar \neg q(a).&\hbox{``Believe that $p(b)$ is false if you believe  $q(a)$ is false''}\\
	\neg q(a).&\hbox{``Believe that $q(a)$ is false''}\\
	\end{array}
	$$
	There is no difference in reasoning about negative literals. In this case, the
only answer set of the program is $\{\neg p(b), \neg q(a)\}$.		
				$\dots$

\smallskip
	\noindent
	{\bf Example 2.2.4.} {\em (Constraints)}\footnote{This example contains a rule with disjunction -- the feature of ASP dialects that we avoid discussing here. Yet, this is an original example by \citeauthor{gelkahl14}~\citeyearpar{gelkahl14} illustrating the role of constraints that take a prominent position in this note.}
	$$
		\begin{array}{ll}
	 p(a)\ or\   p(b). & \hbox{``Believe  $p(a)$ or believe  $p(b)$''}\\
	\ar  p(a). & \hbox{``It is impossible to believe  $p(a)$''}\\
	\end{array}
	$$
	
	The first rule forces us to believe $p(a)$ or to believe $p(b)$. The second rule is
a constraint that prohibits the reasoner’s belief in $p(a)$. Therefore, the first
possibility is eliminated, which leaves $\{p(b)\}$ as the only answer set of the
program. In this example you can see that the constraint limits the sets of
beliefs an agent can have, but does not serve to derive any new information.
Later we show that this is always the case.
				$\dots$

\medskip
	\noindent
	{\bf Example 2.2.5.} {\em (Default Negation)}
	Sometimes agents can make conclusions based on the absence of information. 
For example, an agent might assume that with the absence of evidence
to the contrary, a class has not been canceled. \dots
Such reasoning is captured by default negation.
Here are two examples.
	$$
		\begin{array}{ll}
	 p(a)\ar \naf q(a). & \hbox{``If  $q(a)$ does not belong to your set of  beliefs}\\
	 &  \hbox{then $p(a)$ must''}\\
		\end{array}
	$$
No rule of the program has $q(a)$ in its head, and hence, nothing forces the
reasoner, which uses the program as its knowledge base, to believe $q(a)$.
So, by the rationality principle, he does not. To satisfy the only rule of the
program, the reasoner must believe $p(a)$; thus, $\{p(a)\}$ is the only answer
set of the program. \dots

}
\end{quote}

We now state the informal  semantics hinted by the quoted examples in unifying terms of this paper. 
We denote it by $\infsemGK{\II}$ and
 detail its three components  $\infsemGK{\II}{\structures}$, $\infsemGK{\II}{\formulas}$,  and $\infsemGK{\II}{\models}$. 
 To begin with $\infsemGK{\II}{\structures}$  coincides with 
			$\infsemGL{\II}{\structures}$.
		
		Table~\ref{fig:GKinf:form} presents 	$\infsemGK{\II}{\formulas}$.
			In this presentation 
			we  take the liberty to identify an expression
$
\hbox{{\em (proposition) $p$ does not belong to your set of  beliefs} }
$
used in the examples of the quote listed last  with the expression
$
\hbox{{\em the agent is not made to believe  (proposition) $p$}. }
$
We summarize $\infsemGK{\II}{\models}$  by the entries in  Table~\ref{fig:GLinf:models}, where we replace $\infsemGL{\II}{\models}$ by $\infsemGK{\II}{\models}$.

	\begin{table}
					\caption{The Gelfond-Kahl \citeyearpar{gelkahl14} informal semantics for extended programs with constraints\label{fig:GKinf:form}}
					\begin{center}
						\begin{tabular}{lcp{5.8cm}}
							\hline
							\hspace{2cm}$\Phi$  & \phantom{aa} & \rule{0pt}{0.2cm}\hspace{2cm} $\infsemGK{\II}{\formulas}{\Phi}$\\
							\hline\hline
							propositional atom $A$ & & { believe $\II(A)$ } \rule{0pt}{0.2cm}\\ 
							\hline			
							propositional literal ${\neg}A$ & & believe that $\II(A)$ is false
							\rule{0pt}{0.2cm}\\ 
							\hline 
							expression of the form {\tt not~}$C$ & & the agent is not made to $\infsemGK{\II}{\formulas}{C}$\rule{0pt}{0.2cm}\vspace{0.0cm}\\ 
							\hline 
							expression of the form $\Phi_1,\Phi_2$ & & $\infsemGK{\II}{\formulas}{\Phi_1}$ and $\infsemGK{\II}{\formulas}{\Phi_2}$ \rule{0pt}{0.2cm}\vspace{0.0cm}\\
							\hline
							constraint~~~~~~ $\rul Body$ & & 
							it is impossible to  $\infsemGK{\II}{\formulas}{Body}$\\
							\hline
							rule $Head\rul Body$ & & \begin{minipage}{7cm}
								if $\infsemGK{\II}{\formulas}{Body}$ then $\infsemGK{\II}{\formulas}{Head}$\\
								\emph{(in the sense of material implication)}\end{minipage}\rule{0pt}{0.2cm}\vspace{0.0cm}\\
							\hline
							program $P=\{r_1,\ldots,r_n\}$ & & \begin{minipage}{6cm}
								\rule{0pt}{0.2cm}  All the agent believes is:\\
								$\infsemGK{\II}{\formulas}{r_1}$ and $\infsemGK{\II}{\formulas}{r_2}$  and \dots
								$\infsemGK{\II}{\formulas}{r_n}$
							\end{minipage}\vspace{0.0cm}\\
							\hline\hline
						\end{tabular}
					\end{center}
				\end{table}

\section{Informal semantics of GDT theories  by D-V'12} \label{sec:infsemGDT}
As discussed earlier  \citeauthor{lif02}~\citeyearpar{lif02} coined a term {\gdt} for the commonly used  methodology when applying ASP towards solving difficult combinatorial search problems. Under this  methodology, a program typically consists of three parts: the \gen, \define, and \test groups of rules.

The role of  \gen is to generate the search space. In modern dialects of ASP {\em choice rules} of the form
\beq \{A\} \rul B_1, \dots, B_n, \naf C_1, \dots,\naf C_m,
				\eeq{eq:choice}
are typically used within this part of the program.
Symbols $A$, $B_i$, \cred{and} $C_j$ in~\eqref{eq:choice} are propositional atoms.
The \define part consists of basic rules~\eqref{eq:rule}. This part defines concepts required to state necessary conditions in the \gen and \test parts of the program. The \test part is usually modeled by constraints of the form~\eqref{eq:con},
where $B_i$, \cred{and} $C_j$  are propositional atoms. 

\citeay{den12} defined the logic ASP-FO, where they took the \gen, \define, and \test parts to be the first-class citizens of the formalism. In particular, the ASP-FO language consists of three kinds of expressions G-modules, D-modules, and T-modules. The authors then present formal and informal semantics of the formalism that can be used in practicing ASP. Here we simplify the language ASP-FO by focusing on its propositional counterpart. We call this language GDT. Focusing on the propositional case of ASP-FO helps us in highlighting the key contribution by \citeay{den12} --- the development of {\em objective} informal semantics for logic programs used within ASP or \gdt approach.

A {\em G-module} is a set of choice rules with the same atom in the head; this atom is called {\em open}.  
A {\em D-module} is a  basic logic program whose atoms appearing in the heads of the rules are called {\em defined} or {\em output}.
A {\em T-module} is a constraint.
A {\em GDT theory} is a set of G-modules,  D-modules, and  T-modules so that no G-modules or D-modules coincide on open or defined atoms.
To define the semantics for  {GDT theory} we  introduce several auxiliary concepts including that  of an {\em input answer set}~\citeb{lier11} and {\em G-completion}. 
For a basic program~$\Pi$, we call a set $X$ of atoms an {\em input answer set} of~$\Pi$ if $X$ is an answer set of a program $\Pi\cup (X\setminus{Heads(\Pi)})$, where $Heads(\Pi)$  denotes the set of atoms that occur in the heads of the rules in $\Pi$.

Rules occurring in modules of  {GDT theory} are such that their bodies have the form
\beq
B_1, \dots, B_n, \naf C_1, \dots,\naf C_m.
\eeq{eq:body}
Given $Body$ of the form~\eqref{eq:body} by $Body^{cl}$ we denoted a classical formula of the form
$$B_1\wedge \cdots \wedge  B_n\wedge  \neg C_1 \wedge  \cdots \wedge \neg C_m.$$
For a G-module $G$ of the form $$\{\{A\}\ar Body_1, \dots, \{A\}\ar Body_n\}$$ by {\em G-completion}, $\gcomp(G)$ we denote the  classical formula
$$ A\rightarrow Body_1^{cl}\vee\cdots\vee Body_n^{cl}.$$

For a GDT theory $P$ composed of 
G-modules $G_1,\dots,G_i$, 
D-modules $D_1,\dots,D_j$, 
T-modules $$\rul Body_1,\dots,\rul Body_k,$$ 
we say that  set $X$ of atoms is an {\em answer set} of $\Pi$,
				denoted $X \models_{st} \Pi$, if 
\begin{itemize}[topsep=0pt,itemsep=-1ex,partopsep=1ex,parsep=1ex]
\item $X$ satisfies formulas
$\gcomp(G_1)$, $\dots$, $\gcomp(G_i)$ (we associate a set $X$ of atoms  with an interpretation of classical logic that maps propositional atoms in $X$ to truth value {\em true} and propositional atoms outside of $X$ to truth value {\em false}; we then understand the concept of  satisfaction in usual terms of classical logic.);  
\item $X$ is an input answer set of D-modules $D_1\dots D_j$; and
\item $X$ satisfies formulas $Body_1^{cl}\rightarrow\bot$, $\dots$, $Body_k^{cl}\rightarrow\bot$.
\end{itemize}
We refer the reader to~\citeay{den19} to the discussion of Splitting Theorem results that often allows us to identify ASP logic programs with GDT theories.

					\begin{table}
					\caption{The \citeay{den12} informal semantics for some expressions in GDT theories.\label{fig:DVinf:form}}
					\begin{center}
						\begin{tabular}{lcp{5.8cm}}
							\hline
							\hspace{2cm}$\Phi$  &  & \rule{0pt}{0.2cm}\hspace{2cm} $\infsemDV{\II}{\formulas}{\Phi}$\\
							\hline\hline
							T-theory/constraint~~~ $\rul Body$ & & 
							it is impossible that  $\infsemDV{\II}{\formulas}{Body}$\\
							\hline
							\begin{minipage}{5cm}
							G-module $G$ of the form\\  $\{\{A\}\ar Body_1, \dots, \{A\}\ar Body_n\}$
							\end{minipage}
							& & \begin{minipage}{7cm}
							if $\infsemDV{\II}{\formulas}{A}$ then\\  $\infsemDV{\II}{\formulas}{Body_1}$ or $\dots$ 
							 $\infsemDV{\II}{\formulas}{Body_n}$\\
							 \emph{(in the sense of material implication)}
							 \end{minipage}
							\\
							\hline
							rule $Head\rul Body$ in a D-module& & \begin{minipage}{5.8cm}
								if $\infsemDV{\II}{\formulas}{Body}$ then $\infsemDV{\II}{\formulas}{Head}$\\
								\emph{(in the sense of definitional implication)}\end{minipage}\rule{0pt}{0.2cm}\vspace{0.0cm}\\
							\hline
								\begin{minipage}{5cm}
							D-module $\{r_1,\ldots,r_n\}$ with\\ defined atom $A$ 
								\end{minipage}
							& & \begin{minipage}{5.8cm}
								\rule{0pt}{0.2cm}  All that is known about $A$ is:
								
									$\infsemDV{\II}{\formulas}{r_1}$ and $\infsemDV{\II}{\formulas}{r_2}$  and \dots
								$\infsemDV{\II}{\formulas}{r_n}$
															\end{minipage}\vspace{0.0cm}\\
							\hline
							GDT theory $P=\{M_1,\ldots,M_n\}$& & $\infsemDV{\II}{\formulas}{M_1}$ and $\dots$ $\infsemDV{\II}{\formulas}{M_n}$\\

							\hline\hline
						\end{tabular}
					\end{center}
				\end{table}

We now provide the informal semantics for GDT theory $\Pi$  by~\citeauthor{den12}~\citeyearpar{den12,den19}. 
We denote it by $\infsemDV{\II}$ and
 detail its three components  $\infsemDV{\II}{\formulas}$, $\infsemDV{\II}{\structures}$ and $\infsemDV{\II}{\models}$. 
 To begin with $\infsemDV{\II}{\structures}$  coincides with 
			$\infsemGLorig{\II}{\structures}$. 
			We summarize 	$\infsemDV{\II}{\formulas}$ 
			  by~(i)~the entries in rows 1-3 of Table~\ref{fig:GLinf:form1}, where we replace $\infsemGLorig{\II}{\formulas}$ by $\infsemDV{\II}{\formulas}$, and~(ii)~the entries in
				Table~\ref{fig:DVinf:form}.
			Table~\ref{fig:DVinf:models} presents $\infsemDV{\II}{\models}$. Note how an entry in the right column of Table~\ref{fig:DVinf:models} gives us clues on how to simplify the parallel entry in the right column of Table~\ref{fig:GLinf:models1}. We can rewrite it as follows: {\em For basic program $\Pi$,	property $\infsemGLorig{\II}{\formulas}{\Pi}$ holds in the state $\infsemGLorig{\II}{\structures}{X}$  of affairs}.

								\begin{table}
					\caption{The \citeay{den12} informal semantics 
					for the satisfaction relation.\label{fig:DVinf:models}}
					\begin{center}
					\begin{tabular}{lp{8cm}}
						\hline
						\hspace{0.5cm}$\models_{st}$ &  \hspace{2cm}$\infsemDV{\II}{\models_{st}}$\\
						\hline\hline
						$X\models_{st} \hbox{GDT theory }\Pi$ &  \rule{0pt}{0.4cm}
						Property $\infsemDV{\II}{\formulas}{\Pi}$ holds in the state $\infsemDV{\II}{\structures}{X}$  of affairs.\\
						\hline\hline
					\end{tabular}
					\end{center}
				\end{table}

Provided account of informal semantics of GDT theories
echos the interpretation of an answer set of a basic program as a possible ``interpretation'' and can be seen as an informal semantics for the syntactic constructs that are fundamental in ASP practice nowadays.

\section{Conclusions and Acknowledgments}
In this note, we reviewed four papers and their accounts on informal semantics of logic programs under answer set semantics. We put these accounts into a uniform perspective by focusing on three components of each of the considered informal semantics, namely, (i) the interpretation of answer sets;  (ii) the interpretation of syntactic expressions; and
(iii)  the interpretation of semantic satisfaction relation. We also discussed the relations of the presented informal semantics to two programming paradigms that emerged in the field of logic programming after the inception of the concept of a stable model: ASP and \aspp.

We would like to thank Michael Gelfond, Marc Denecker, Jorge Fandinno, Vladimir Lifschitz, Miroslaw Truszczynski, Joost Vennekens for fruitful discussions on the topic of this note. Marc Denecker brought my attention to the subject of informal semantics and his enthusiasm for the questions pertaining to this subject was contagious.  

The author was partially supported by NSF 1707371.
			
\bibliographystyle{tlplike}	
\bibliography{asp-fo}
\end{document}